\documentclass{article}

\usepackage{microtype}
\usepackage{graphicx}
\usepackage{subfigure}
\usepackage{booktabs}
\usepackage{enumitem}
\usepackage{multirow}
\usepackage{balance}
\usepackage{bbm, bm}
\usepackage{amsmath}
\usepackage{amsfonts}
\usepackage{hyperref}

\usepackage[accepted]{icml2021}

\DeclareMathOperator*{\argmin}{arg\,min}

\begin{document}

\twocolumn[
\icmltitle{Estimating the Uncertainty of Neural Network Forecasts for Influenza Prevalence Using Web Search Activity}

\begin{icmlauthorlist}
\icmlauthor{Michael Morris}{ucl}
\icmlauthor{Peter Hayes}{ucl}
\icmlauthor{Ingemar J. Cox}{ucl,diku}
\icmlauthor{Vasileios Lampos}{ucl}
\end{icmlauthorlist}
\icmlaffiliation{ucl}{Department of Computer Science, University College London, UK}
\icmlaffiliation{diku}{Department of Computer Science, University of Copenhagen, Denmark}
\icmlcorrespondingauthor{Michael Morris}{michael.morris.19@ucl.ac.uk}
\icmlcorrespondingauthor{Peter Hayes}{peter.hayes.15@ucl.ac.uk}
\icmlcorrespondingauthor{Ingemar J. Cox}{i.cox@ucl.ac.uk}
\icmlcorrespondingauthor{Vasileios Lampos}{v.lampos@ucl.ac.uk}
\icmlkeywords{uncertainty, forecasting, Bayesian neural networks, influenza, web search}
\vskip 0.3in
]

\printAffiliationsAndNotice{}

\begin{abstract}
Influenza is an infectious disease with the potential to become a pandemic, and hence, forecasting its prevalence is an important undertaking for planning an effective response. Research has found that web search activity can be used to improve influenza models. Neural networks (NN) can provide state-of-the-art forecasting accuracy but do not commonly incorporate uncertainty in their estimates, something essential for using them effectively during decision making. In this paper, we demonstrate how Bayesian Neural Networks (BNNs) can be used to both provide a forecast and a corresponding uncertainty without significant loss in forecasting accuracy compared to traditional NNs. Our method accounts for two sources of uncertainty: data and model uncertainty, arising due to measurement noise and model specification, respectively. Experiments are conducted using $14$ years of data for England, assessing the model's accuracy over the last $4$ flu seasons in this dataset. We evaluate the performance of different models including competitive baselines with conventional metrics as well as error functions that incorporate uncertainty estimates. Our empirical analysis indicates that considering both sources of uncertainty simultaneously is superior to considering either one separately. We also show that a BNN with recurrent layers that models both sources of uncertainty yields superior accuracy for these metrics for forecasting horizons greater than $7$ days.
\end{abstract}

\section{Introduction}
Influenza is an infectious respiratory disease responsible for $290{,}000$ to $650{,}000$ deaths annually according to the World Health Organisation. Estimating the prevalence of influenza-like-illness (ILI) in a population and forecasting its future trajectory is an important area of research~\cite{shaman2012forecasting, shaman2013real, yang2015forecasting, nsoesie2014systematic} as it can contribute to better-informed policy decisions for public health, e.g. when to recommend the use of antiviral drugs. Established approaches to epidemic forecasting, such as mechanistic models, are commonly based on syndromic, clinical, and demographic data~\cite{birrell2011Bayesian,shaman2012forecasting}. These methods attempt to mathematically model a disease, usually as a set of ordinary differential equations. They rely on assumptions such as random and uniform mixing of populations which often do not hold \cite{roberts2015nine}. In recent years, a considerable body of work has shown that data generated by web users during their interaction with search engines or social media platforms can be used to obtain influenza prevalence models with sufficient accuracy. This has been shown both for \emph{now}-casting~\cite{lampos2010a,culotta2010towards,lampos2015advances,yang2015accurate,wagner2018added} and forecasting~\cite{paul2014twitter,volkova2017forecasting,yang2017using}.

Neural Networks (NN) have been applied to many forecasting tasks with state of the art accuracy, including financial modelling \cite{gately1995neural}, hydrological forecasting \cite{thirumalaiah2000hydrological}, power load forecasting \cite{barbounis2006long}, as well as to the task of influenza forecasting from social media data \cite{volkova2017forecasting}. However, traditional NNs do not provide uncertainty estimates. Influenza forecasts, and forecasts in general, are only useful when they have an associated uncertainty. It is problematic to make an informed decision based on a forecast without knowing the likelihood that it is accurate. Uncertainty estimates also aid the fusion of disparate information sources, which are common in syndromic surveillance of diseases.

There are generally two main types of uncertainty in modelling problems \cite{der2009aleatory}: data uncertainty and model uncertainty. Model uncertainty, also known as \emph{epistemic}, deals with uncertainty in the model's parameters \cite{kendall2017uncertainties}. This captures uncertainty about information that is not contained in the training dataset. It can be reduced by having a more extensive dataset; a model will be more confident about data that is very similar to what it has seen during training, and less confident about out-of-distribution data. This is relevant to our task as disease transmission patterns and user search behaviour are expected to change over time. Bayesian models place a prior distribution over the model parameters \cite{yao2019quality, li2017dropout, foong2019between} to account for model uncertainty, but they neglect data uncertainty \cite{gal2016uncertainty}. Data uncertainty, also known as \emph{aleatoric}, is inherent in the observations and is often caused by noise or sampling errors. This is relevant to our task as search query data can be extremely noisy (e.g. data represents a pseudo-random 10-15\% sample of all searches) and the uncertainty in the ILI rate changes throughout the year with very little uncertainty outside the main circulation months. Data uncertainty is modelled by placing a distribution at the output of an NN~\cite{bishop1994mixture, nix1994estimating, le2005heteroscedastic}. 

To account for both of these uncertainties we combine the methods, creating a Bayesian Neural Network (BNN) which has an output distribution as described in \cite{kendall2017uncertainties}. We find that a combination of these two methods gives a good estimation of uncertainty without compromising forecasting performance when forecast horizon windows are greater than $7$ days. 

Our paper makes the following contributions:
\begin{enumerate}[topsep=0pt, partopsep=0pt, leftmargin=12.5pt]
\setlength{\itemsep}{2pt}
\setlength{\parskip}{0pt}
    \item We propose an uncertainty modelling solution for ILI forecasting using web search data and BNN layers. We show that our method can be incorporated into common NN architectures, such as feed-forward (FF) and recurrent neural networks (RNN).
    \item We investigate the performance on ILI forecasting of traditional FF and RNN models and compare it to models that account for model or data uncertainty, as well as their combination.
    \item In addition to common regression error functions (e.g. mean squared error), we also use metrics which penalise both errors in the mean and confidence interval of a forecast. We show that estimating uncertainty causes little or no accuracy degradation, and that an RNN that simultaneously estimates model and data uncertainty provides the most accurate forecasts and confidence intervals.
\end{enumerate}

\section{Related Work}
Methods for time series forecasting are applicable to many different areas, from meteorology~\cite{murphy1984probability,gneiting2005weather} to financial modelling~\cite{kaastra1996designing,cao2001financial, abu1996introduction}, and health~\cite{brookmeyer2007forecasting,hoot2008forecasting,soyiri2013overview}. 

Traditional NNs have shown promise as forecasters for influenza and are well-suited for incorporating large feature spaces that capture aggregate online user activity~\cite{aiken2019towards, volkova2017forecasting, venna2018novel, xue2017influenza, adhikari2019epideep}. However, their original formulations do not provide confidence intervals for forecasts, something that reduces their practical utility. This has also been highlighted by influenza forecasting challenges such as CDC's FluSight~\cite{reich2019collaborative}. Gaussian Processes (GP) have been used to \emph{now}-cast or forecast influenza or COVID-19 mortality with uncertainty estimates \cite{lampos2015advances,zimmer2020influenza,lampos2021covid}. However, scalability to larger datasets that from a modelling standpoint cover more meaningful time-spans, is challenging as GPs have $\mathcal{O}(n^3)$ complexity where $n$ is the number of training samples~\cite{liu2020gaussian}. NNs can be modified to provide similar uncertainty estimates. Dropout can be used as an approximation to deep GP models~\cite{gal2016dropout, gal2016theoretically, kendall2015Bayesian, gal2016uncertainty}. This requires careful tuning and ultimately does not behave as desired given that the uncertainty generated by dropout does not reduce as more data becomes available~\cite{osband2016risk, hron2017variational}. Another approach is to use BNNs which specify a distribution over their weights and attempt to learn the posterior distribution. The majority of BNNs in forecasting are optimised using Markov Chain Monte Carlo (MCMC)~\cite{liang2005Bayesian, niu2012short, cauchemez2004Bayesian, li2011Bayesian, zhang2011explicitly}, but are limited by the lack of scalability of existing MCMC algorithms for large training samples and feature spaces~\cite{papamarkou2019challenges,kucukelbir2017automatic}. This prevents MCMC from being a viable solution for neural network ILI forecasting with uncertainty estimates. Variational approaches that fit an approximate posterior can avoid the scaling issues of MCMC~\cite{hinton1993keeping, blundell2015weight}. However, variational methods have been rarely adopted as they tend to deliver non-optimal performance, usually converging very slowly~\cite{graves2011practical, blundell2015weight}. Recent works applying the heuristics of batch normalisation and learning rate scheduling appear to have reduced this problem~\cite{ioffe2015batch, loshchilov2016sgdr, goyal2017accurate, osawa2019practical}. 

In this paper, we use variational inference~\cite{blei2017variational} to optimise a BNN for model uncertainty. We also combine this with data uncertainty by outputting a distribution, computing uncertainties as in \citeauthor{kendall2017uncertainties}~(\citeyear{kendall2017uncertainties}). This improves the final uncertainty estimates and is computationally tractable.

\section{Methods}

\subsection{Problem Formulation}
For the forecasting task, we have a set of inputs $\mathbf{X} = \left[\mathbf{x}_1, \dots, \mathbf{x}_T \right]$ and a set of target outputs $\mathbf{y} = \left[y_{1}, \dots, y_T\right]$ across $T$ time points (days). 
At each time point $t \in [1,T]$, we construct an input $\mathbf{x}_t$ composed of $m$ web search query frequencies from time $t-\ell$ to time $t$, and ILI rates from $t-\ell-\delta$ to $t-\delta$. The number of historical time points that we consider is denoted by $\ell$ and $\delta$ is a delay (in days) due to the time it takes for public health systems to obtain and curate a representative sample of ILI rates based on doctor visitations. The input $\mathbf{x}_t$ is an $(m+1)\times\ell$ matrix which is flattened into an $(m+1)\times\ell$-dimensional vector when used in an FF NN; it is maintained as an $(m+1)\times\ell$ matrix when an RNN is used. We also have a target ILI rate $y_t$, associated with $\mathbf{x}_t$, which is a future ILI rate at time point $t+\gamma$, where $\gamma$ is the forecasting horizon. In our experiments, we conduct $\gamma = 7$, $14$, and $21$ days ahead forecasting. For simplicity, in the remainder of the manuscript, we will drop the subscript $t$ and refer to these variables as $\mathbf{x}$ and $y$.

\subsection{Neural Network Architectures for ILI Rate Forecasting}
We experiment both with FF and RNN NN models. For the RNN we use the Long Short-Term Memory (LSTM) architecture. We obtain forecasts using deterministic model formulations, models that account for data uncertainty, BNNs that account for model uncertainty, and BNNs that combine model and data uncertainty.

\textbf{Feed-Forward Model.} A single layer FF NN is used as the simplest model we modify with these uncertainty methods. We flatten\footnote{Flattening refers to turning a matrix into a vector in a column-major order.} each input $\mathbf{x}$ and pass it into a hidden dense layer with $25$ units and a ReLu activation function $(\max(0,x))$. This goes into the output layer which is dependent on the uncertainty method we employ and will be described in the following paragraphs. All FF models are trained with an exponential learning rate scheduler using ADAM as an optimiser~\cite{kingma2014adam}.

\textbf{Long Short-Term Memory Model.} RNNs are well-suited and common for time series modelling tasks \cite{kalchbrenner2013recurrent, sundermeyer2012lstm}. LSTMs~\cite{LSTM} are a popular form of an RNN. We use many-to-one LSTM architectures \cite{graves2006connectionist} only predicting the ILI rate at a single time point $\gamma$ days ahead. Input data is passed to the model sequentially using a rolling window. As the main focus of our work is in incorporating uncertainty estimates and not in assessing complex NN architectures, we deploy the following straightforward architecture. We use a single LSTM layer that returns only the final sequence of predictions, followed by a dense layer, and then an output layer. More complex architectures need to be investigated with the caveat that our dataset is still relatively small when it comes to NNs (\(14\) years of daily data) and optimising a network with many parameters will be more challenging in this case. The output layers are again dependent on the uncertainty method we employ. All LSTM models are trained with a cosine learning rate scheduler and an ADAM optimiser.

\subsection{Modelling Uncertainty}
Traditional NNs are deterministic and require modification to provide estimates of uncertainty. We first present the deterministic approach without any uncertainty consideration, and then we detail our approach to data or model uncertainty, as well as their combination. 

\textbf{Deterministic Approach.} Let us assume that an NN with $L$ layers is being deployed. In our experiments, $L=1$ or $2$ for the FF- or the LSTM-based NN architectures, respectively. We denote the \(j^{\text{th}}\) activation in the \(l^{th}\) layer as \(a_j^{[l]}\). The output layer has a single unit and generates a point prediction $\hat{y} = f_{\mathbf{\Phi}}(\mathbf{x})$ where $f_\mathbf{\Phi}$ denotes an NN with parameters composed of weights and biases $\mathbf{\Phi}$. We optimise $\mathbf{\Phi}$ by minimising the mean squared error (MSE) between \(\mathbf{\hat{y}}\) and \(\mathbf{y}\) for \(T\) time points. We use backpropagation to find values for $\mathbf{\Phi}$.

\textbf{Data Uncertainty.} Data uncertainty is caused by noise in the data. This usually arises due to sampling or measurement error. It can be represented as homoscedastic uncertainty where uncertainty is constant for every input, or as heteroscedastic uncertainty where the uncertainty is dynamic and dependent on input \(\mathbf{x}\). Heteroscedastic models are useful where parts of the observation space may be noisier than others \cite{gal2016uncertainty}. The ground truth ILI rate is nearly constant when influenza is not circulating, but highly variable during the flu season. This makes a homoscedastic model unsuitable for this problem. As such we only consider the heteroscedastic case. 

We assume that the uncertainty is normally distributed for each data point. We separate our prediction into a mean and standard deviation \cite{kendall2017uncertainties}:
\begin{equation}
    \label{data_uncertainty_prediction}
    \left[\hat{y}, \hat{\sigma}\right] = f_{\mathbf{\Phi}}(\mathbf{x})\, .
\end{equation}
To accomplish this, we expand the output layer of the NN from one to two units \(\left[a^{[L]}_1, a^{[L]}_2\right]\)~\cite{bishop1994mixture}. The mean, $\hat{y}$, is the first activation of the output layer that uses a linear activation function
$\hat{y}=a^{[L]}_1$.
The standard deviation, \(\hat{\sigma}\), is computed by taking the softplus of \(a^{[L]}_2\):
\begin{equation}
    \hat{\sigma} = \frac{1}{\rho}\ln({1+e^{\rho a^{[L]}_2}}) \, ,
    \label{softplus_eq}
\end{equation}
where \(\rho > 0\) is a sharpening factor (see Appendix). The softplus ensures that the standard deviation is always positive. As we are assuming that the noise is normally distributed and we are outputting the parameters of a Gaussian distribution, we train the model using maximum likelihood estimation. This is the same as minimising the negative-log-likelihood (NLL) :
\begin{equation}
\small
\label{NLL_eq}
\text{NLL}(\mathbf{y}, \mathbf{\hat{y}}, \mathbf{\hat{\bm{\sigma}}}) 
= \frac{1}{T}\sum_{t=1:T}^{T}\left( 
\frac{1}{2 \hat{\sigma}_t^2}{(y_t-\hat{y}_t)}^2
+\frac{1}{2}\log{(2 \pi \hat{\sigma}_t^2)} \right) \, ,
\end{equation}
where $\mathbf{y}=(y_1,\dots,y_T)$ is a series of ILI rates (ground truth), $\hat{\mathbf{y}}=(\hat{y}_1,\dots,\hat{y}_T)$ is a series of ILI rate predictions, and $\hat{\bm{\sigma}}=(\hat{\sigma}_1,\dots,\hat{\sigma}_T)$ is a series of associated uncertainties. We use NLL as a loss function and use backpropagation to find values for $\mathbf{\Phi}$. The first component of Eq.~\ref{NLL_eq} contains a residual term equivalent to MSE and an uncertainty normalisation term. The second component prevents the model from predicting an infinitely large $\sigma$. We do not need to explicitly train with an `uncertainty ground truth' as it is implicit in the NLL.

\textbf{Model Uncertainty.} Model uncertainty is caused by the model having a limited understanding of the system which generates the data \cite{dicing_with_the_unknown, tagasovska2019single}. This uncertainty can usually be reduced by observing more data. An architecture which accounts for model uncertainty should recognise when it is shown out-of-distribution inputs and be less confident. This is not possible while using data uncertainty alone.

To capture model uncertainty a prior distribution is first placed over the model parameters, e.g. $\mathbf{\Phi} \sim \mathcal{N}(0,I)$ for a Gaussian prior. In the literature this is referred to as a BNN~\cite{li2017dropout, yao2019quality, foong2019between}. Bayesian inference is used to compute the posterior distribution over the parameters \(p(\mathbf{\Phi}|\mathbf{X}, \mathbf{y})\). We can then sample from this distribution $K$ times and make $K$ predictions to build a range of forecasts. The variance of the means of these forecasts is the variance due to model uncertainty. For each sample from the posterior a prediction is made, $p(y|f_{\mathbf{\Phi^\prime}}(\mathbf{x})) = \mathcal{N}\left({f_{\mathbf{\Phi^\prime}}(\mathbf{x})}, \sigma \right)$, where $f_{\mathbf{\Phi^\prime}}(\mathbf{x})$ is calculated by the model from a sample of weights and $\sigma$ is a hyper-parameter. Each prediction is required to be a distribution so that the expected likelihood for each sample, $\mathbf{\Phi^\prime}$, can be calculated during training. The scale of $\sigma$ is set to be similar to $\mathbf{y}$. 

BNNs are difficult to perform inference on. The posterior is given by 
\begin{equation}
\label{eq:int2}
        p(\mathbf{\Phi}|\mathbf{X}, \mathbf{y})=
        \frac{ p(\mathbf{y}|\mathbf{X},\mathbf{\Phi})p(\mathbf{\Phi})} 
        {p(\mathbf{y}|\mathbf{X})}\, .
\end{equation}
The denominator of Eq.~\ref{eq:int2} contains the marginal probability $p(\mathbf{y}|\mathbf{X})$ which cannot be computed analytically in this setting \cite{kendall2017uncertainties}. To mitigate against this several approximations have been proposed~\cite{graves2011practical, blundell2015weight, gal2016dropout, blei2017variational}. In approximate inference the averaging over all weights is replaced by an optimisation task. We constrain the posterior \(p(\mathbf{\Phi}|\mathbf{X}, \mathbf{y})\) to a simple distribution \(q_{\mathcal{Q}}(\mathbf{\Phi})\), where $\mathcal{Q}$ is a family of potential distributions which we define as Gaussians. We choose a \(q(\mathbf{\Phi})\) which minimises the Kullback-Leibler (KL) divergence to the true posterior \(p(\mathbf{\Phi}|\mathbf{X}, \mathbf{y})\):
\begin{equation}
\label{kl_eq}
    \argmin_{q(\mathbf{\Phi})\in \mathcal{Q}} D_{\text{KL}}\left[q(\mathbf{\Phi})||p(\mathbf{\Phi}|\mathbf{X}, \mathbf{y})\right] \, .
\end{equation}
This is also intractable as it contains the marginal probability from Eq.~\ref{eq:int2}. We can, however, avoid this issue by minimising the negative evidence-lower-bound (ELBO) given by:
\begin{equation}
\small
    \label{elbo_eq}\text{ELBO}(\omega, \theta)= \mathbb{E}\left[\log\left(p(\mathbf{X},\mathbf{y}|\mathbf{\Phi})\right)\right]-D_{\text{KL}}\left[q_{\theta}(\mathbf{\Phi})||p_{\omega}(\mathbf{\Phi})\right] \, , 
\end{equation}
where $\omega$ and $\theta$ are parameters used to learn $p_{\omega}(\mathbf{\Phi})$ and $q_\theta(\mathbf{\Phi})$ (described later). This is equal to Eq.~\ref{kl_eq} up to a constant \cite{blei2017variational}. The first component of Eq.~\ref{elbo_eq} is the expected likelihood that encourages the model to choose a \(q_{\theta}(\mathbf{\Phi})\) which explains the data well. The second component is the negative KL divergence between the posterior and prior distribution. This behaves similarly to a regulariser and encourages the model to choose a simple $q_{\theta}(\mathbf{\Phi})$. The prior is chosen to represent how we anticipate the model to behave before it has observed any data. When there is limited training data, the prior component dominates the loss function. When more data is observed, the model becomes more confident and relies more on the likelihood component. 

The prior and posterior distributions are parameterised by functions that are conditioned on $\mathbf{x}$ and output the parameters of a Gaussian distribution. These functions are single-layer FF NNs with parameters $\omega$ and $\theta$, respectively. In the prior we fix the standard deviation and parameterise the mean, $p_{\omega}(\mathbf{\Phi})=\mathcal{N}(\mu_p, \sigma_p) = \mathcal{N}\left(f_{\omega}(\mathbf{x}), \sigma_p \right)$. In the posterior we parameterise both the mean and standard deviation, $q_{\theta}(\mathbf{\Phi}) = \mathcal{N}\left(\mu_q, \sigma_q\right) = \mathcal{N}\left(f_{\theta}\left(\mathbf{x}\right)\right)$. Here $f_{\theta}\left(\mathbf{x}\right)$ outputs a mean $\mu_q$ and standard deviation $\sigma_q$ in the same way as in Eq.~\ref{data_uncertainty_prediction} (using $\rho_q$ as a sharpening factor).

In practise, we utilise a BNN with a posterior distribution over the parameters in the final layer only. The other layers of the network can be interpreted as providing a lower dimensional representation of the inputs to a Bayesian model \cite{tran2018bayesian}. During training we sample once from $q_{\theta}(\Phi)$ and compute the ELBO. We take gradients with respect to $\omega$ and $\theta$ to perform gradient descent and optimise the Bayesian layer. We back-propagate the loss through the sampled weights to optimise deterministic weights in the preceding network layers. At prediction time, the model is called $K$ times. Weights are sampled $\mathbf{\Phi}^\prime \sim q(\mathbf{\Phi})$ each time the model is called and thus give a different prediction $\mathcal{N}(\hat{y}^\prime, \sigma)$. We assume that these predictions are normally distributed and use them to calculate the parameters of a Gaussian distribution $\mathcal{N}(\hat{y}, \hat{\sigma})$, where \(\hat{y}\) and $\hat{\sigma}$ are the mean and standard deviation of the means $\hat{y}^\prime$ of $K$ sampled forecasts. As we are using a Gaussian output and only our output layer uses a distribution over its weights, it would be possible to analytically compute the posterior distribution. However, we choose to use variational inference for this task as it is a more general solution that would allow any layer in the network to use a distribution over its weights~\cite{blei2017variational}.

\textbf{Combining Model and Data Uncertainty.} To capture both data and model uncertainty we combine the data uncertainty model with a BNN. To do this we use a BNN which outputs \(\left[\hat{y},\hat{\sigma}\right]\). We approximate the posterior over the BNN by minimising the negative ELBO, using $\hat{\sigma}$ in the computation of the expected likelihood. At prediction time we draw weights from our posterior $\mathbf{\Phi}^\prime \sim q_{\theta}(\mathbf{\Phi})$ and use this to obtain a model output with predictive mean and standard deviation $\left[\hat{y}^\prime, \hat{\sigma}^\prime\right] = f_{\mathbf{\Phi^\prime}}(\mathbf{x})$. This process is repeated $K$ times, each giving a different prediction. The mean of our combined uncertainty prediction is the mean of the  $K$ sampled forecasts, $\hat{y} = \mu(\mathbf{\hat{y}^\prime})$.
The standard deviation of the predicted distributions is given by \cite{kendall2017uncertainties}:
\begin{equation}
\label{combine_uncertainties_eq}
    \hat{\sigma} \approx
    \sqrt{\frac{1}{K} \sum^K_{\kappa=1} \hat{\mathbf{y}}^{\prime 2}_\kappa
    - \left(\frac{1}{K} \sum^K_{\kappa=1} \hat{\mathbf{y}}^\prime _\kappa\right)^2
    + \frac{1}{K} \sum^K_{\kappa=1} \hat{\bf{\sigma}}^{\prime 2}_\kappa}\,.
\end{equation}

\subsection{Error Metrics}
We evaluate forecasting accuracy based on the following metrics: mean absolute error (MAE), root mean squared error (RMSE), bivariate correlation ($r$), and symmetric mean absolute percentage of error (SMAPE) between the predicted, $\hat{\mathbf{y}} = [\hat{y}_1,\dots,\hat{y}_T]$, and the target, $\mathbf{y} = [y_1,\dots,y_T]$, values. SMAPE accounts for different magnitudes in different flu seasons and provides more relevant error estimates when averaging across all the test periods. We also use a smoothed delay-to-peak (SDP) metric, which measures the delay between the actual and the predicted peak of a flu season in days, after smoothing both time series ($15$-day moving average). Smoothing is relevant to avoid misleading outcomes when, for example, a flu season has two peaks in close temporal proximity. A negative SDP value means that the model predicted the peak early, while a positive one means the peak was predicted late. We consider its absolute value when computing the mean performance over multiple test seasons.

\setlength{\tabcolsep}{2.5pt}
\begin{table*}[t!]
\caption{Metrics averaged across 4 test years for \texttt{FF}, \texttt{LSTM}, and $3$ baseline models. We evaluate the statistical significance between the metrics within a model architecture. We use $^\star$ and $\dagger$ superscripts to indicate that an estimate is different in a statistically significant way from the deterministic baseline of the same architecture and  an equivalent model in the other NN architecture, respectively. The best result for each metric is in bold.  We append model abbreviations with \texttt{-v} for (vanilla) deterministic models, \texttt{-d}, \texttt{-m}, and \texttt{-c} for data, model, and combined uncertainty, and \texttt{-c-nq} for combined uncertainty without using search queries.}
\label{tab:performance_main}
\vskip 0.1in
\begin{center}
\small
\begin{tabular}{ll|ccc|ccccc|ccccc}
\toprule
              &                & \multicolumn{3}{|c|}{\bf Baselines}               & \multicolumn{5}{|c|}{\bf \texttt{FF}}                                                         & \multicolumn{5}{|c}{\bf \texttt{LSTM}} \\
$\bm{\gamma}$ & \textbf{Error} & \bf Na\"ive & \bf Hist. & \bf \texttt{GP} & \bf \texttt{-v}  & \bf \texttt{-d} & \bf \texttt{-m} &   \bf \texttt{-c} & \bf \texttt{-c-nq} & \bf \texttt{-v}   & \bf \texttt{-d}   & \bf \texttt{-m}   & \bf \texttt{-c}   & \bf \texttt{-c-nq}  \\
\midrule
7     &    CRPS &          -- &   $2.62$ &    $2.16$ &                -- &           $2.44$ &                    $1.6$ &  $\mathbf{1.54}$ &           $2.42$ &                 -- &            $2.0$ &           $1.89^{\dagger}$ &                           $1.72$ &             $1.7^{\dagger}$ \\
      &     NLL &          -- &   $3.03$ &   $29.28$ &                -- &           $5.59$ &                   $2.73$ &  $\mathbf{2.19}$ &           $2.75$ &                 -- &           $3.82$ &           $3.64^{\dagger}$ &                           $2.26$ &            $2.27^{\dagger}$ \\
      &     MAE &      $2.33$ &   $3.12$ &     $2.5$ &   $\mathbf{1.62}$ &   $3.26^{\star}$ &           $2.13^{\star}$ &   $2.12^{\star}$ &   $3.22^{\star}$ &   $2.29^{\dagger}$ &           $2.77$ &           $2.35^{\dagger}$ &                           $2.29$ &            $2.12^{\dagger}$ \\
      &    RMSE &      $3.74$ &   $5.18$ &    $4.66$ &    $\mathbf{2.5}$ &   $4.98^{\star}$ &           $3.32^{\star}$ &   $3.37^{\star}$ &   $4.94^{\star}$ &   $3.84^{\dagger}$ &   $4.84^{\star}$ &           $3.97^{\dagger}$ &                           $3.82$ &            $3.41^{\dagger}$ \\
      &   SMAPE &     $13.45$ &  $17.26$ &   $12.83$ &  $\mathbf{11.55}$ &  $22.32^{\star}$ &          $15.26^{\star}$ &  $14.42^{\star}$ &  $24.16^{\star}$ &            $13.31$ &          $15.21$ &                    $14.82$ &                          $14.21$ &           $14.54^{\dagger}$ \\
      &     $r$ &       $0.9$ &   $0.83$ &    $0.88$ &   $\mathbf{0.95}$ &   $0.87^{\star}$ &           $0.92^{\star}$ &           $0.94$ &   $0.85^{\star}$ &             $0.92$ &           $0.89$ &           $0.89^{\dagger}$ &                 $0.91^{\dagger}$ &            $0.91^{\dagger}$ \\
      &     SDP &     $-14.0$ &  $21.25$ &     $9.5$ &            $9.11$ &          $20.89$ &  $\mathbf{5.39^{\star}}$ &           $9.75$ &  $17.89^{\star}$ &  $13.78^{\dagger}$ &          $21.36$ &          $12.06^{\dagger}$ &                          $12.69$ &                     $16.06$ \\
\midrule
14    &    CRPS &          -- &   $2.62$ &    $2.91$ &                -- &           $2.63$ &                   $2.08$ &           $2.08$ &           $2.64$ &                 -- &           $2.48$ &                     $2.11$ &                  $\mathbf{1.86}$ &            $2.09^{\dagger}$ \\
      &     NLL &          -- &   $3.03$ &   $43.99$ &                -- &           $6.73$ &                    $4.0$ &           $2.71$ &           $2.91$ &                 -- &           $6.26$ &                     $3.59$ &                           $2.46$ &   $\mathbf{2.44^{\dagger}}$ \\
      &     MAE &      $3.23$ &   $3.12$ &    $3.24$ &   $\mathbf{2.45}$ &   $3.44^{\star}$ &                   $2.65$ &   $2.81^{\star}$ &   $3.44^{\star}$ &             $2.54$ &   $3.29^{\star}$ &                     $2.63$ &                           $2.51$ &            $2.81^{\dagger}$ \\
      &    RMSE &      $5.06$ &   $5.18$ &    $5.88$ &   $\mathbf{3.73}$ &    $5.4^{\star}$ &                   $3.94$ &   $4.41^{\star}$ &   $5.07^{\star}$ &             $4.16$ &   $5.64^{\star}$ &                     $4.23$ &                           $4.39$ &            $4.55^{\dagger}$ \\
      &   SMAPE &      $17.9$ &  $17.26$ &   $16.27$ &           $16.32$ &          $20.91$ &                  $18.14$ &  $18.98^{\star}$ &  $25.71^{\star}$ &            $14.86$ &  $17.35^{\star}$ &                     $16.4$ &        $\mathbf{14.1^{\dagger}}$ &      $17.24^{\dagger\star}$ \\
      &     $r$ &      $0.82$ &   $0.83$ &     $0.8$ &            $0.92$ &           $0.88$ &           $0.88^{\star}$ &  $\mathbf{0.92}$ &   $0.85^{\star}$ &              $0.9$ &            $0.8$ &                      $0.9$ &                  $0.9^{\dagger}$ &       $0.87^{\dagger\star}$ \\
      &     SDP &     $-21.0$ &  $21.25$ &   $19.25$ &           $14.22$ &          $26.14$ &          $\mathbf{10.5}$ &          $12.31$ &          $17.69$ &            $17.08$ &          $25.22$ &          $17.19^{\dagger}$ &                          $14.31$ &           $14.58^{\dagger}$ \\
\midrule
21    &    CRPS &          -- &   $2.62$ &    $3.66$ &                -- &           $2.66$ &                   $3.09$ &            $2.4$ &           $3.12$ &                 -- &           $2.62$ &           $2.56^{\dagger}$ &                   $\mathbf{2.2}$ &            $2.44^{\dagger}$ \\
      &     NLL &          -- &   $3.03$ &   $54.38$ &                -- &           $6.99$ &                    $9.9$ &            $3.5$ &           $3.17$ &                 -- &          $10.42$ &            $5.3^{\dagger}$ &                           $2.88$ &   $\mathbf{2.64^{\dagger}}$ \\
      &     MAE &      $4.05$ &   $3.12$ &    $4.26$ &            $3.11$ &           $3.45$ &           $3.66^{\star}$ &           $3.14$ &   $3.87^{\star}$ &             $3.17$ &           $3.26$ &           $3.07^{\dagger}$ &                  $\mathbf{2.89}$ &            $3.24^{\dagger}$ \\
      &    RMSE &      $6.24$ &   $5.18$ &     $6.9$ &   $\mathbf{4.76}$ &           $5.37$ &            $5.6^{\star}$ &           $4.93$ &   $5.44^{\star}$ &    $5.2^{\dagger}$ &           $5.63$ &           $4.94^{\dagger}$ &                           $4.93$ &                       $5.1$ \\
      &   SMAPE &     $22.12$ &  $17.26$ &    $25.7$ &            $21.1$ &          $21.56$ &                  $22.89$ &          $20.58$ &  $28.36^{\star}$ &  $17.93^{\dagger}$ &          $16.78$ &          $19.44^{\dagger}$ &  $\mathbf{15.96^{\dagger\star}}$ &      $20.36^{\dagger\star}$ \\
      &     $r$ &      $0.72$ &   $0.83$ &    $0.72$ &            $0.85$ &           $0.84$ &           $0.78^{\star}$ &           $0.87$ &           $0.83$ &             $0.85$ &           $0.84$ &  $\mathbf{0.88^{\dagger}}$ &                           $0.87$ &                      $0.83$ \\
      &     SDP &     $-28.0$ &  $21.25$ &    $17.5$ &           $22.17$ &          $16.47$ &          $16.64^{\star}$ &          $16.03$ &  $14.06^{\star}$ &            $24.94$ &          $22.97$ &                    $19.81$ &                          $20.47$ &     $\mathbf{11.0^{\star}}$ \\
\bottomrule
\end{tabular}
\end{center}
\vskip -0.1in
\end{table*}

For probabilistic forecasts, the aforementioned metrics cannot distinguish between errors with large or small uncertainties. We need an error metric that penalises both the forecast and associated uncertainty estimates. NLL, used to train the NNs which we defined in Eq.~\ref{NLL_eq} is one such metric. A criticism of NLL is that it over-penalises errors where the difference between the actual and forecasted value is much greater than the associated uncertainty \cite{gneiting2007strictly}. An alternative that does not exhibit this property is the continuous ranked probability score (CRPS)~\cite{gneiting2007strictly} which is given by 
\begin{multline}
    \text{CRPS}(\mathbf{y}, \mathbf{\hat{y}},\mathbf{\hat{\sigma}}) = \frac{1}{T} \sum^T_{t=1}\hat{\sigma}  \Bigg[ \frac{1}{\sqrt{\pi}}-2\varphi_t \left(\frac{y_t-\hat{y}_t}{\hat{\sigma}_t} \right) \dots \\
    - \frac{y_t-\hat{y}_t}{\hat{\sigma}_t} \left( 2\Psi_t \left(\frac{y_t-\hat{y}_t}{\hat{\sigma}_t} \right) -1\right)\Bigg] \, ,
\end{multline}
where \(\varphi_t\) and \(\Psi_t\) denote the probability density function and the cumulative distribution function of a standard Gaussian variable $\mathcal{N}(\hat{y}_t,\hat{\sigma}_t)$. CRPS is a probabilistic metric that generalises to the MAE when the standard deviation is \(0\). A discussion of the trade-off between NLL and CRPS is provided in the Appendix.

\begin{figure*}[t!]
    \centering
        \includegraphics[width=0.99\textwidth]{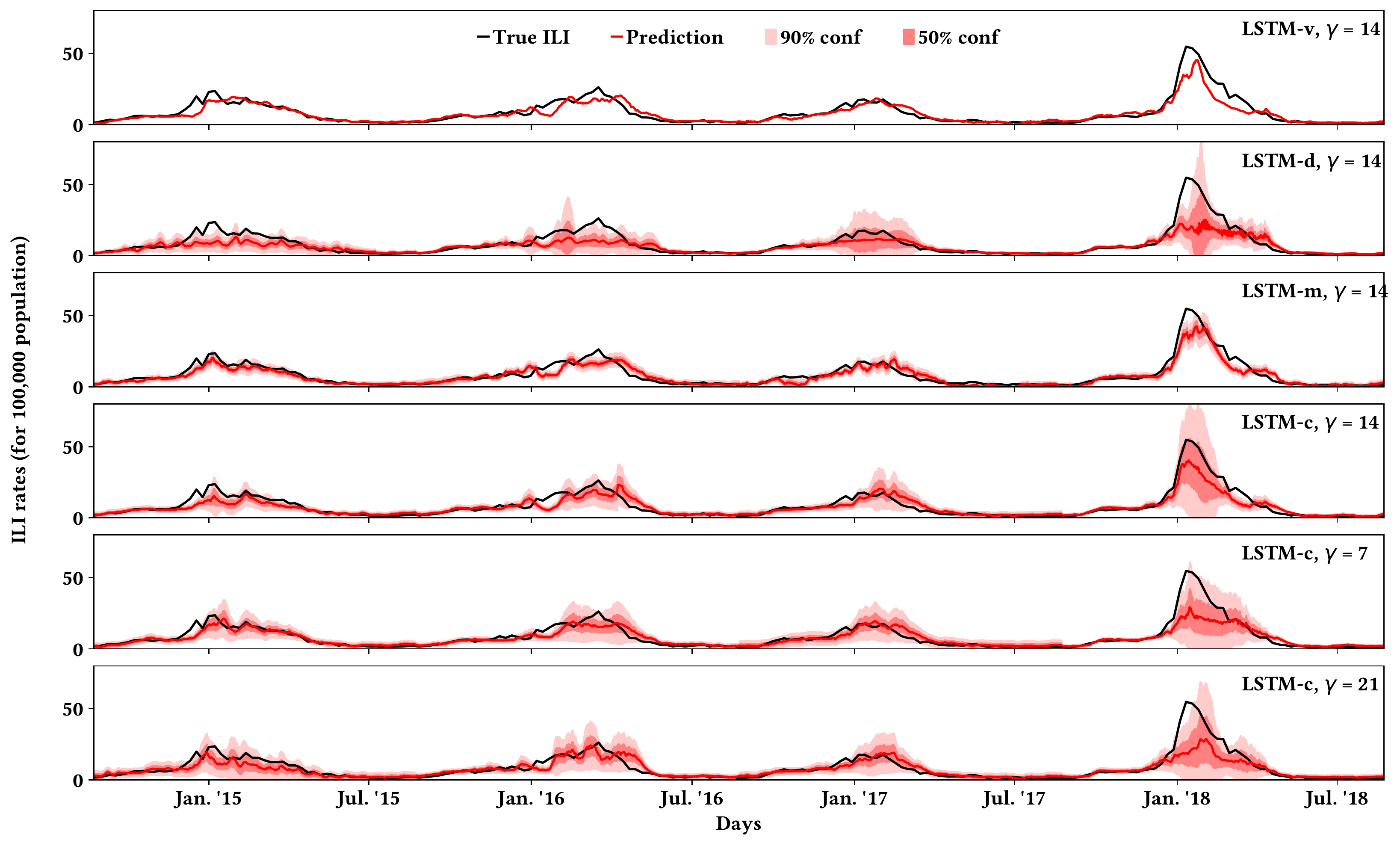}
        \caption{Comparison of LSTM models for ILI forecasting tasks. \texttt{LSTM-v} is a deterministic model, and \texttt{LSTM-d}, \texttt{LSTM-m}, \texttt{LSTM-c} estimate data, model, and combined uncertainty, respectively. The number following the model names denotes $\gamma$ (the forecasting horizon). The corresponding figure with FF models is in the Appendix.}
        \label{main_results_fig}
\end{figure*}

\begin{figure}[t!]
\centering

\includegraphics[width=0.99\columnwidth]{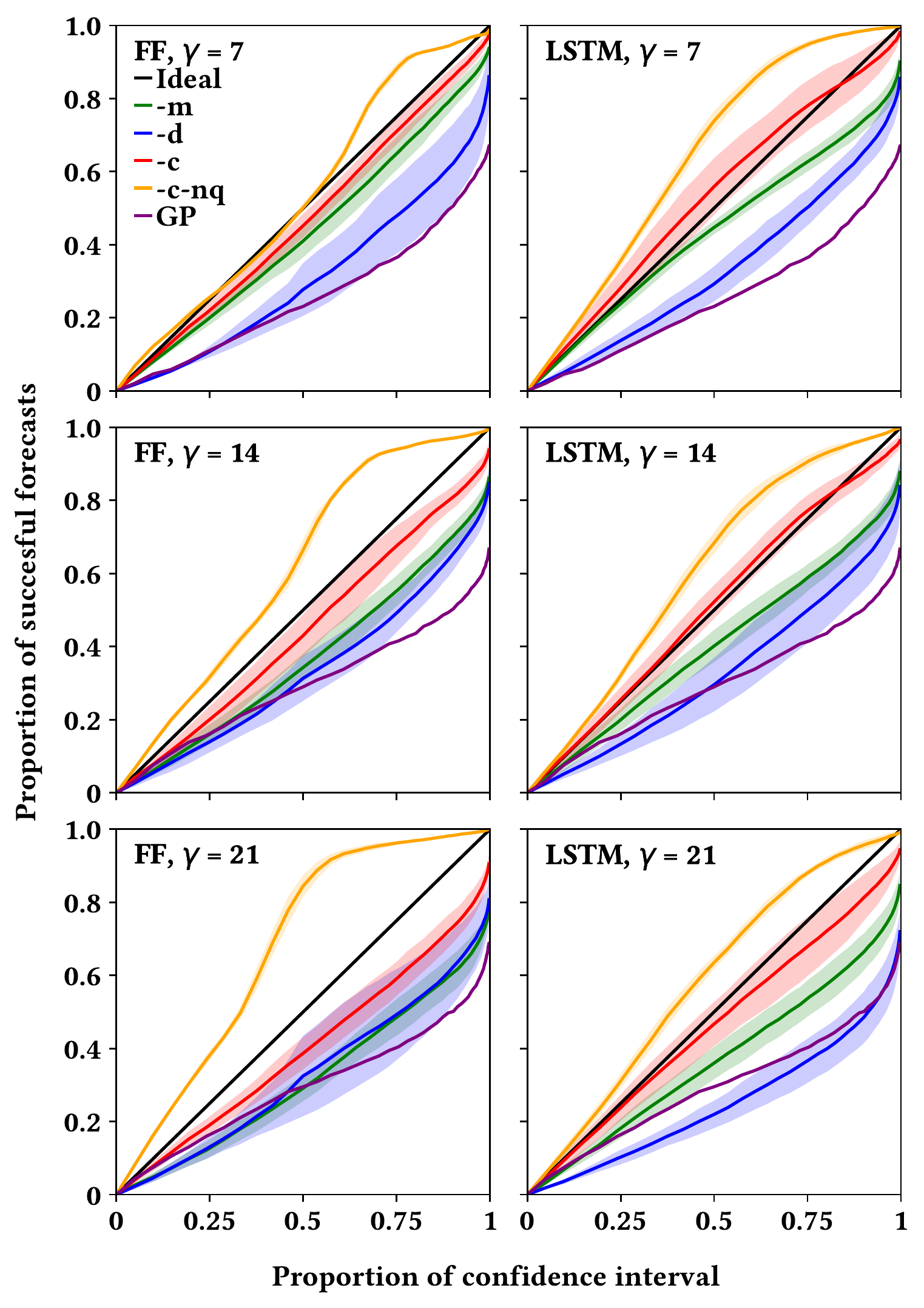}
\caption{Charts showing uncertainty calibration of various models. The x-axis shows the predicted probability of a $\gamma$ days ahead forecast, whereas the y-axis shows how often the ground truth falls within this range. \texttt{-v} denotes vanilla deterministic models, \texttt{-d}, \texttt{-m}, and \texttt{-c} data, model, and combined uncertainty, and \texttt{-c-nq} combined uncertainty without using search queries. Confidence intervals indicate the standard deviation of the mean obtained by training with different weight initialisation seeds over $10$ runs.}
\label{calibration_fig}
\end{figure}

\section{Results and analysis}

\subsection{Datasets}
Weekly ILI rates for England from January 1, 2004 to December 30, 2018 are obtained by the Royal College of General Practitioners (RCGP), which utilises a sentinel doctor network spread across the country. Weekly ILI rates represent the proportion of patients (in $100{,}000$ of the population) visiting medical clinics with symptoms of influenza. Weekly data is converted to daily through a linear interpolation centred around Thursday of each week. Daily web search frequencies for England are obtained from the Google Health Trends API for the same period as the ILI rates. We use a predetermined pool of $20{,}856$ health-related search queries. We apply a semantic filter based on word embedding representations to extract queries that are related to influenza \cite{Lampos2017www,Bill_GP,Zou2019}. We also filter queries based on their correlation with ILI rates. The selected queries change with the test year as the correlations can differ from one training period to another. On average across the $4$ test periods, $170$ queries are selected.\footnote{Access to the Google Health Trends API is controlled by Google. Application to obtain access is open to all researchers.} We also smooth query frequencies using a harmonic weighted average over the past $7$ days.

\subsection{Experiment Settings}
\textbf{Evaluation Protocol.} We report results averaged over 4 test periods (2014-15 to 2017-18) that cover the respective annual flu seasons. Each test period starts on August 23 and ends on August 22 of the following year. Min-max normalisation is applied to each search query frequency time series (based on the training set each time). We do not normalise the ILI rate inputs as they are on the same scale as the output, and this normalisation can negatively affect forecasting accuracy. In all models, the number of historical time points, $\ell$, is 28 days for both search queries and ILI rates. ILI rates are delayed by a week, i.e. $\delta=7$ to simulate a practical setting -- syndromic surveillance data is commonly reported with a delay. We report results for $\gamma = 7$, $14$, and $21$ days.

\textbf{Baselines.} We compare our approach to a GP model as formulated by~\citeauthor{zimmer2020influenza}~(\citeyear{zimmer2020influenza}). This model is trained using the latest available data at the start of each week. The GP formulation does not use search query data -- an increased input space dimensionality needs a different covariance structure otherwise it negatively affects accuracy. To compare with it in a more meaningful way we also train models that do not use web search activity data. In addition to the NN architectures, we also include a na\"ive model (also known as a persistence model) as a benchmark. This uses the most recently available value of the input ILI rate as its forecast. Finally, a historical averages model (Hist.) computes the ILI rate and uncertainty on a date as the average and variance of ILI rates on the same date in previous years. The na\"ive estimates are superior for short-term forecasts while historical averages are better for long-term forecasts. 

\subsection{Results}
\label{sec:results}

Table~\ref{tab:performance_main} enumerates performance outcomes for all forecasting tasks, models, and error metrics we considered, averaged over the $4$ year-long test periods. For all metrics except the correlation ($r$) a lower number is better. We append NN model abbreviations with \texttt{-v} for (vanilla) deterministic models, \texttt{-d}, \texttt{-m} or \texttt{-c} respectively for data, model or combined uncertainty, and \texttt{-c-nq} for combined uncertainty without using search query data. We find that combined \texttt{-c} models generally perform the best in terms of their confidence intervals and accuracy, however, there is an exception for forecasting $7$ days ahead ($\gamma = 7$). Here the deterministic \texttt{FF-v} model outperforms the combined \texttt{FF-c} model by $21\%$ and $33\%$ in terms of SMAPE and RMSE, respectively. This is in part due to the the model's good performance on the 2017-18 flu season which had a higher peak. Arguably for decision and planning purposes, precise uncertainty intervals may be an acceptable compromise for a small accuracy reduction. We find that for forecast horizons, $\gamma=14$ or $21$ days, the \texttt{LSTM-c} model is equivalent or better than all other methods. We also evaluate the effect of using search query frequencies as inputs in \texttt{FF-c} and \texttt{LSTM-c} models, by removing the queries from the inputs and training otherwise identical models. We find that models trained without search query data are less accurate and less confident, confirming that search query data is useful for forecasting.
 
\textbf{Standard Regression Accuracy Metrics.} Standard metrics (MAE, RMSE, SMAPE, and $r$) do not consider corresponding uncertainty estimates, but are, of course, important. NNs are superior to the baseline models with the exception of data uncertainty models \texttt{-d}. For forecasting horizon $\gamma=7$, \texttt{FF} models provide the best forecasts. For $\gamma = 14$, the \texttt{FF-v} and \texttt{LSTM-v} are not statistically different, and \texttt{-v}, \texttt{-m} and \texttt{-c} models perform similarly. For $\gamma = 21$, which is the most challenging forecasting task in our experiments, the best-performing models are based on the LSTM architecture. We see that for $\gamma=7$ or $14$ days, the \texttt{LSTM-c} model is not different in a statistically significant way from the baseline \texttt{LSTM-v} model. However, for $\gamma=21$ this is not the case; according to SMAPE, \texttt{LSTM-c} is the best performing model in a statistically significant way.

\textbf{Uncertainty.} The combined uncertainty \texttt{-c} models always perform better than data uncertainty \texttt{-d} and model uncertainty \texttt{-m} models in terms of both NLL and CRPS. Data or model uncertainty when used in isolation tend to underestimate uncertainty; this is resolved when used together. We find that the \texttt{-c-nq} models sometimes outperform the \texttt{-c} models in terms of NLL. This is due to the NLL's tendency to heavily penalise forecasts which are over-confident. The \texttt{-c-nq} models are by far the least confident of the models (Figures~\ref{FF_nq_plot}~and~\ref{LSTM_no_queries_fig}) and as a result, are less likely to have a bad prediction which strongly affects NLL. Figure~\ref{main_results_fig} shows how the uncertainty differs between different models. The \texttt{-m} model underestimates uncertainty but has a reasonably accurate mean, while the \texttt{-d} model has an inaccurate mean and less confidence. The \texttt{-c} model has a good mean prediction and a reasonable confidence interval (metrics for this are shown in Table~\ref{tab:performance_main}). Figure~\ref{main_results_fig} also shows that as $\gamma$ increases, the uncertainty surrounding the forecasts increases as expected.

We evaluate how well the uncertainties are calibrated in the same way as in~\citeauthor{kendall2017uncertainties}~(\citeyear{kendall2017uncertainties}). We generate confidence intervals for $0$ to $3\sigma$ confidence ($0\%$ to $99.7\%$) and compute the frequency that the ground truth falls within this forecast. For example, we expect the ground truth to be within a $50\%$ confidence interval $50\%$ of the time. We show this in calibration plots in Figure~\ref{calibration_fig}. The diagonal line $y = x$ represents perfect calibration. When a model is too confident, it will be below this line, and when it is too uncertain, it will be above it. In every case the combined models have the best calibrated uncertainty, i.e. their curves are closest to $y=x$. \texttt{LSTM-c} and \texttt{FF-c} have the best-calibrated uncertainty for all $\gamma$'s. We also observe that as $\gamma$ increases, the calibration lines deviate further from the diagonal. The GP model underestimates uncertainty significantly. Finally, \texttt{-c-nq} models overestimate uncertainty, which is also reflected in their low NLL values.

\textbf{Epidemic Forecasting Analysis.}
During the flu season 2015-16, the ILI rate peaked much later than in other seasons (mid-March compared to early January in most seasons). As a result, models tend to overestimate the flu rate in January and then predict the true peak at an inaccurate future point, resulting in a much larger SDP. The peak intensity of the flu season does not appear to affect the SDP. Flu rates in 2017-18 were significantly increased compared to other recent flu seasons, so the models tended to underestimate the flu rate. In addition, the larger face value of the maximum flu rate during this season increases the error metrics (other than the SDP) for it compared to other years. The models are most accurate during the 2014-15 and 2016-17 flu seasons. In general, the variation in intensity and timing during the flu season make forecasting challenging. Because of this, the models must exhibit meaningful and well-tuned uncertainty estimates which reflect the variations year to year. In our experiments, the models with combined uncertainty are visibly the best ones in their attempt to capture this (Figures~\ref{main_results_fig}~and~\ref{calibration_fig}).

\section{Conclusion}
The uncertainty in a forecast is important to decision making and planning, especially in the context of public health interventions. We showed how BNNs can be used to model both data and model uncertainties. Empirical assessment showed that for a $7$ day forecast horizon there was some degradation in forecast accuracy when uncertainty is considered. However, this knowledge of uncertainty may be a worthwhile compromise for a small accuracy reduction. For longer time horizons ($14$, $21$ days) there was little or no degradation in performance while still having accurate estimates of uncertainty. Modelling both uncertainties proved to be better than either method individually. This was confirmed by calibration curves (Figure \ref{calibration_fig}). LSTM models were best for $14$ and $21$ days ahead. The ability to make probabilistic forecasts with NNs makes them a viable option for real-world epidemiological forecasting where knowledge of uncertainty is required.

\section*{Acknowledgements}
The authors would like to thank Simon Moura for the constructive feedback in early versions of this work. We would also like to acknowledge RCGP for providing syndromic surveillance data, and Google for providing access to the Google Health Trends API. I.J.C. and V.L. would like to acknowledge all levels of support from the EPSRC project ``i-sense: EPSRC IRC in Agile Early Warning Sensing Systems for Infectious Diseases and Antimicrobial Resistance'' (EP/R00529X/1). I.J.C. and V.L. would also like to acknowledge the support from a Google donation funding the project ``Modelling the prevalence and understanding the impact of COVID-19 using web search data''. 

\bibliography{refs}
\bibliographystyle{icml2021}

\icmltitlerunning{Estimating the Uncertainty of Neural Network Forecasts for Influenza Prevalence Using Web Search Activity}

\section*{Appendix}

\textbf{Trade-off between CRPS and NLL.} Here we discuss the differences between the CRPS and NLL loss functions. The optimal solution to CRPS is the same as for NLL (e.g. confident and correct). CRPS, however, is more forgiving when the confidence is high and the accuracy is poor. Figure~\ref{nll_crps} illustrates this point. Here the blue and green curves depict the NLL and CRPS scores, respectively, as a function of the variance (x-axis). Predictions are represented by the red diagonal line. The true value to be predicted is $y=0$. The first point on the diagonal line has zero variance, but predicts $y=-1$, i.e. an erroneous value with perfect confidence (zero uncertainty). We observe that the CRPS penalises this with a score of $1$. In contrast the NLL tends to infinity. As we move from left to right, the error in $y$ is initially decreasing while our uncertainty is increasing. As our estimate approaches the true value of $y=0$ which occurs when the standard deviation (x-axis) is $0.25$, both curves approach a minimum value. As we continue to move from left to right, the error in $y$ begins to increase along with the uncertainty. At the right-most side, we have $y=1$ with a standard deviation of $0.5$. Here, the CRPS score is similar to when the standard deviation was zero, while the NLL is about $2.3$, i.e. the NLL metric much more strongly penalises errors that are outside of the uncertainty region. We do not favour one measure over the other, and report both in our results.

\renewcommand{\thefigure}{A\arabic{figure}}
\setcounter{figure}{0}
\begin{figure}[ht]
\centering
\includegraphics[width=0.99\columnwidth]{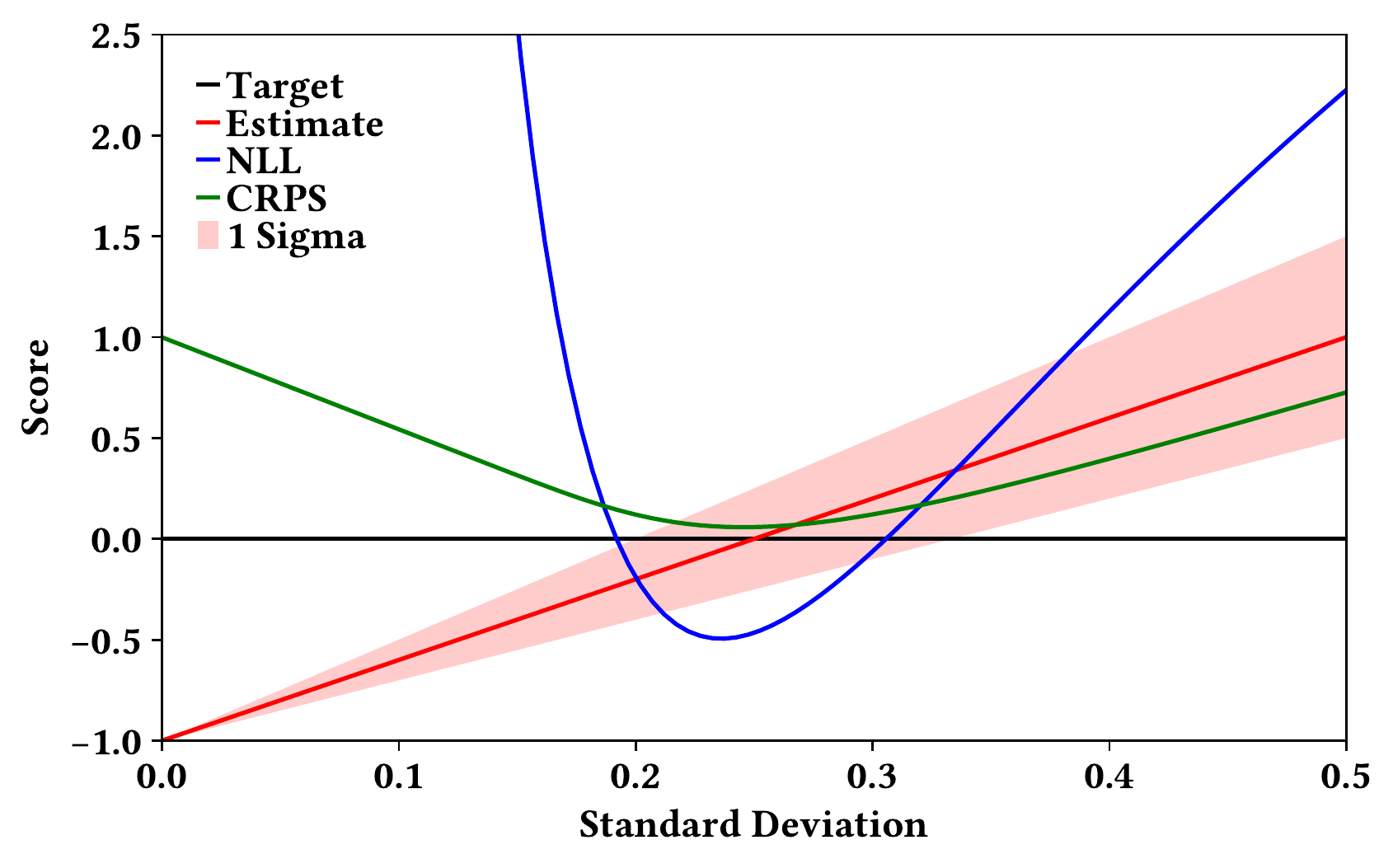}
\caption{NLL and CRPS variation with changing error and confidence. The red line shows a model's prediction with standard deviation shown in pink where the true value is $y=0$. As the accuracy and confidence change, the CRPS and NLL values have different trajectories.}
\label{nll_crps}
\end{figure}

\textbf{Hyper-Parameter Settings.}
We have not optimised hyper-parameters using a validation set as our main focus was not in establishing state-of-the-art performance. Our results are still significantly better compared to their respective baselines. In our experiments, we have used the following hyper-parameter settings. For the $\texttt{-d}$ NNs, we set $\rho = 0.25$. For the \texttt{-m} NNs, we set $\rho_q = 10$ (posterior sharpening factor), $\sigma_p = 0.5$ (prior standard deviation), and $\sigma = 5$ (output standard deviation). For the \texttt{-c} NNs, we set $\rho_q = 10$, $\sigma_p = 0.5$, and $\rho = 0.25$. We use batch normalisation to improve the training of all models besides the deterministic FF model that is negatively affected by it. Batch normalisation layers using default parameters (momentum$=0.99$, epsilon$=0.001$) are inserted between layers of the network, and normalise the input's mean and variance. In the models that combine uncertainty types, we investigated the effect that the number $K$ of sampled models has on the prediction. We looked at values for $K$ between $2$ and $1{,}000$ and found that for $K > 25$ the evaluation metrics do not change significantly. Therefore, in our experiments, we use $K=100$ as this ensures that our output distribution converges. We found that for FF models an exponential learning rate scheduler is best. LSTM models are trained with a cosine learning rate scheduler with warm-up \cite{loshchilov2016sgdr, goyal2017accurate}. We train the models for a fixed amount of $200$ epochs.

\textbf{Statistical Significance.} We use a two-tailed $t$-test with Bonferroni correction to evaluate if there is a statistically significant ($p \leq 0.05$) change in accuracy between different approaches. We train each model $10$ times with different initialisation seeds and generate metrics for each for comparison.

\textbf{Limitations.} The models in this work are limited in that they are dependent on the existence of search query and epidemiological data. This data is only available in countries where both a significant Internet usage and an established health infrastructure are present. The ground truth is also not representative of the true ILI prevalence in the population as it is measured by the RCGP and does not account for people who do not go to the doctor when they have ILI symptoms. However, we expect that any consistent biases in the ground truth will not affect the relevance of the machine learning task. Finally, we have not optimised the hyper-parameters of the NNs, but just set them manually to some reasonable values. The performance we obtained was still significantly better than the baselines we compared it against.

\textbf{Future Work.} Future work will look at improving the accuracy of the forecasting models by incorporating more complex architectures. Solutions might also explore the use of probabilistic weights in more layers of the network to give more accurate posterior approximations. Assessing the accuracy of our approach in more countries and sub-regions as well as on different infectious diseases, e.g. COVID-19, are also natural next steps.

\begin{figure*}[!ht]
\centering
\includegraphics[width=0.99\textwidth]{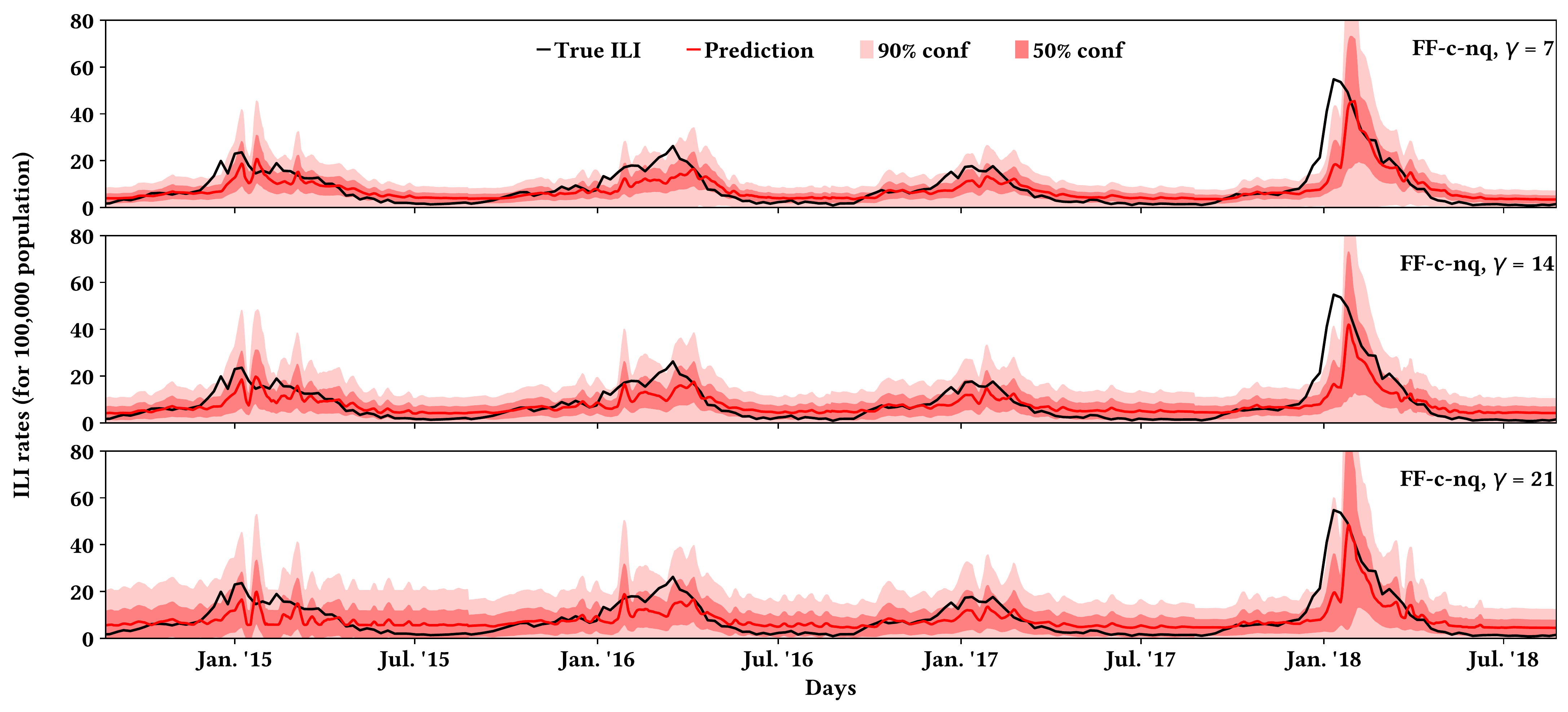}
\caption{Forecast estimates for $\gamma = 7$, $14$, $21$ for the \texttt{FF-c-nq} model.}
\label{FF_nq_plot}
\end{figure*}

\begin{figure*}[!ht]
\centering
\includegraphics[width=0.99\textwidth]{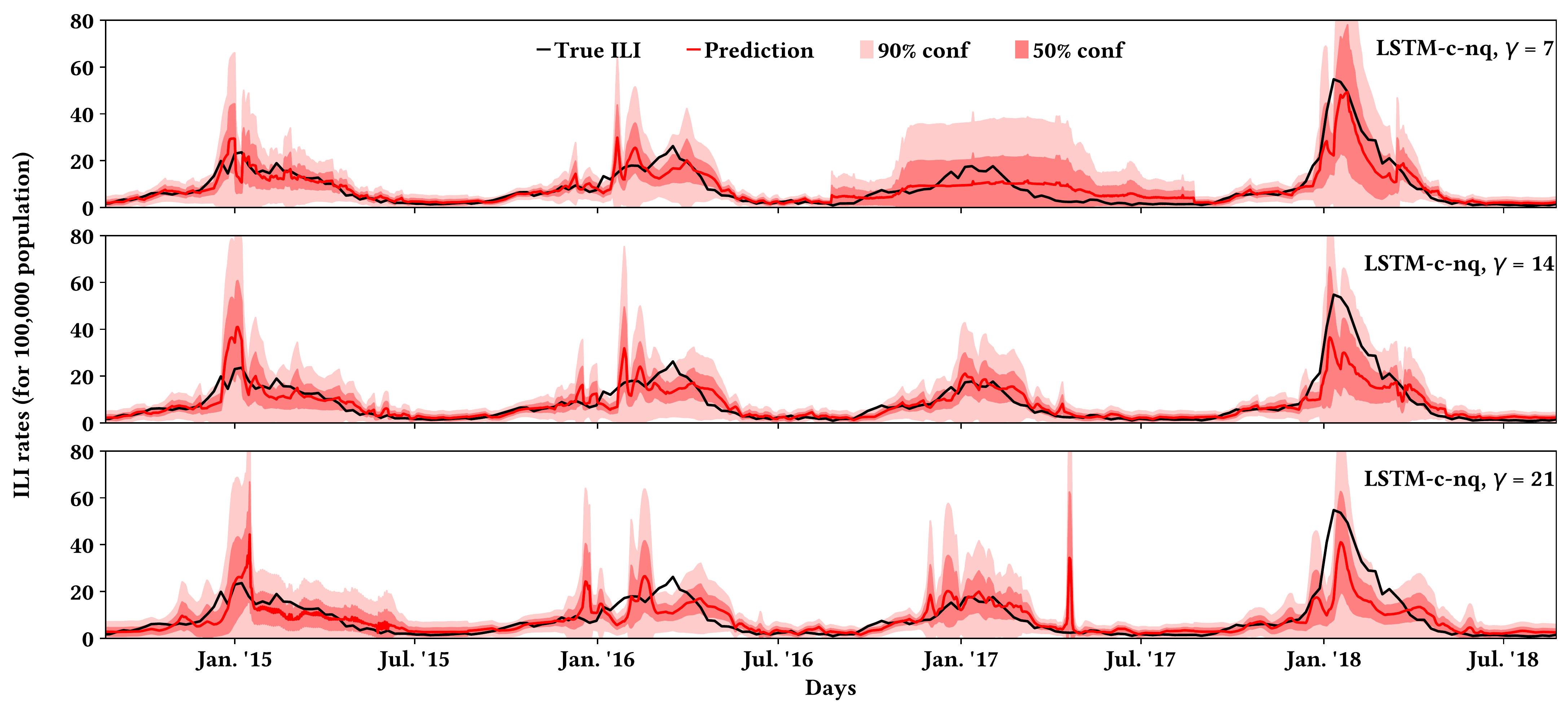}
\caption{Forecast estimates for $\gamma = 7$, $14$, $21$ for the \texttt{LSTM-c-nq} model.}
\label{LSTM_no_queries_fig}
\end{figure*}

\begin{figure*}[t]
\centering
\includegraphics[width=0.99\textwidth]{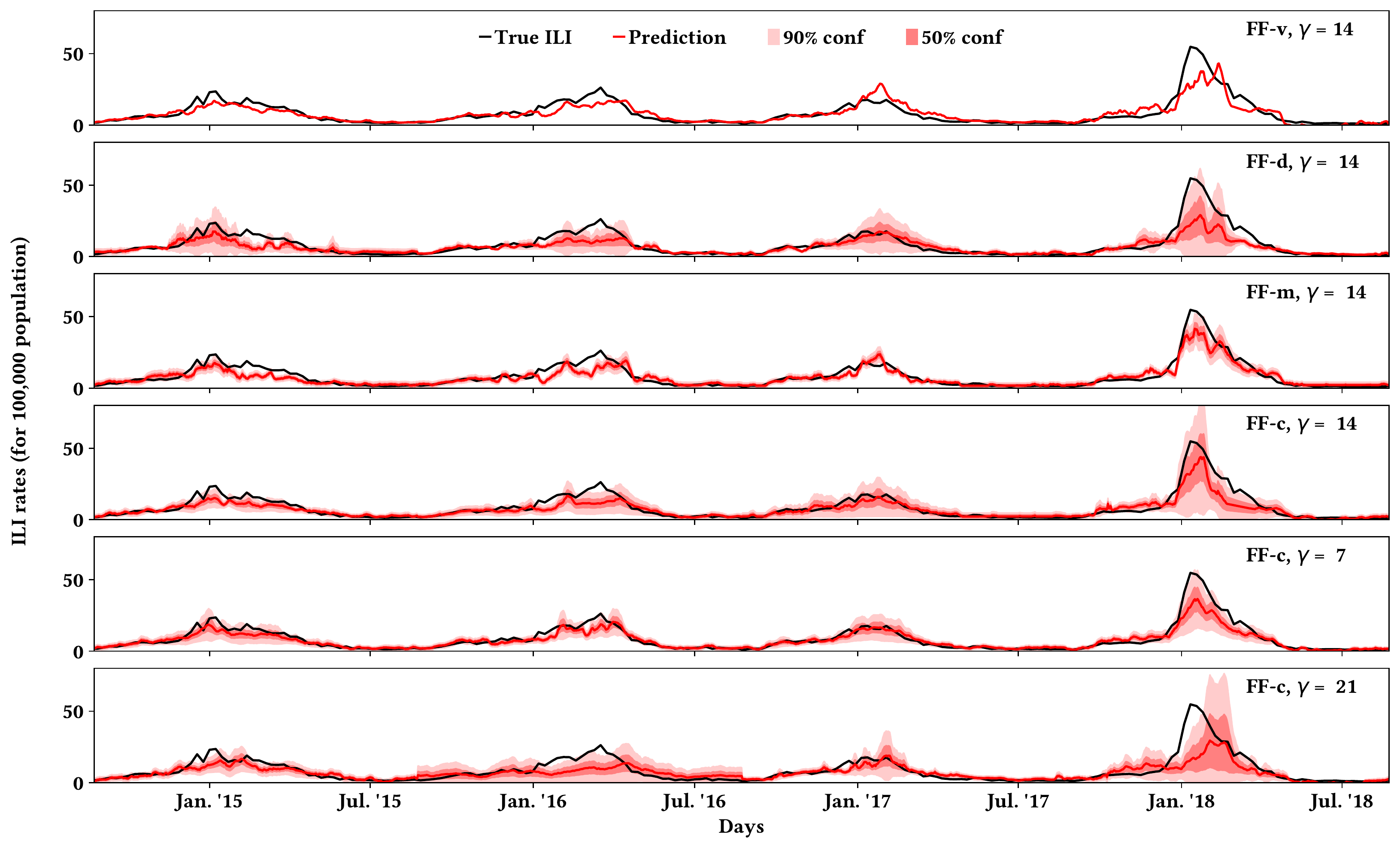}
\caption{
Comparison of FF models for ILI forecasting tasks. \texttt{FF-v} is a deterministic model, and \texttt{FF-d}, \texttt{FF-m}, \texttt{FF-c} estimate data, model, and combined uncertainty, respectively. The number following the model names denotes $\gamma$ (the forecasting horizon). It can be seen that as $\gamma$ increases the predictions become less accurate and less confident.}
\label{FF_main_plot}
\end{figure*}

\textbf{Additional Figures.} We show figures for both \texttt{FF-c-nq} and \texttt{LSTM-c-nq}. In Figures~\ref{FF_nq_plot}~and~~\ref{LSTM_no_queries_fig}, we can see that the models that do not use web search data are not yielding confident estimates for any of the considered forecasting horizons. The predictions from these models contain many invalid estimates (incorrect spikes and multiple peaks during a single flu season). As expected, this issue becomes more pronounced for greater forecasting horizons. Notably, the forecasts for the 2017-18 flu season are surprisingly good for both models. This can cause metrics which are dependent on scale to be skewed against models which perform poorly on this season. To this end, the SMAPE scores for models without queries are much poorer. In Figure~\ref{FF_main_plot}, we show the predictions obtained from the FF models. The accuracy for $\gamma=7$ is better than the one obtained from LSTM models, however for $\gamma=14$ and $21$ the LSTM models are comparable or better (see Figure~\ref{main_results_fig}).

\end{document}